\documentclass[runningheads]{llncs}
\usepackage{graphicx}
\usepackage{listings}
\usepackage{xspace}
\usepackage{algorithm}
\usepackage{multirow}
\usepackage{color}
\usepackage{colortbl}
\usepackage{tabularx}
\usepackage{multirow}
\usepackage{subfigure}
\usepackage{eqparbox}
\usepackage{booktabs}
\usepackage[english]{babel}
\usepackage{xcolor}
\usepackage{svg}
\usepackage{listings}
\usepackage{algpseudocode}
\usepackage{amsmath}

\newcolumntype{b}{X}
\newcolumntype{s}{>{\hsize=.45\hsize}X}

\newcommand\cparagraph[1]{\vspace{1mm}\noindent\textbf{#1}\xspace}

\definecolor{Gray}{gray}{0.95}
\definecolor{highlight}{rgb}{1,1,0.04}

\let\OLDthebibliography\thebibliography
\renewcommand\thebibliography[1]{
  \OLDthebibliography{#1}
  \setlength{\parskip}{0pt}
  \setlength{\itemsep}{0pt}
}

\begin{document}

\title{Optimizing Sparse Matrix Multiplications for Graph Neural Networks}

\author{Shenghao Qiu\inst{1} \and
Liang You\inst{2} \and Zheng Wang\inst{1}\thanks{This project was supported in part by an Alibaba Innovative Research Programme.
 For any correspondence, please contact Zheng Wang (E-mail:
z.wang5@leeds.ac.uk).}}
\institute{University of Leeds, Leeds, UK\\
\email{\{sc19sq, z.wang5\}@leeds.ac.uk}\\
\and
Alibaba Group, Beijing, China\\
\email{youliang.yl@alibaba-Inc.com}}

\maketitle

\vspace{-2mm}
\begin{abstract}
Graph neural networks (GNNs) are emerging as a powerful technique for modeling graph structures. Due to the sparsity of real-world graph
data, GNN performance is limited by extensive sparse matrix multiplication (SpMM) operations involved in computation. While the right
sparse matrix storage format varies across input data, existing deep learning frameworks employ a single, static storage format, leaving
much room for improvement. This paper investigates how the choice of sparse matrix storage formats affect the GNN performance. We observe
that choosing a suitable sparse matrix storage format can significantly improve the GNN training performance, but the right format depends
on the input workloads and can change as the GNN iterates over the input graph. We then develop a predictive model to dynamically choose a
sparse matrix storage format to be used by a GNN layer based on the input matrices. Our model is first trained offline using training
matrix samples, and the trained model can be applied to any input matrix and GNN kernels with SpMM computation. We implement our approach
on top of PyTorch and apply it to 5 representative GNN models running on a multi-core CPU using real-life and synthetic datasets.
Experimental results show that our approach gives an average speedup of 1.17x (up to 3x) for GNN running time.

\end{abstract}

\vspace{-6mm}
\section{Introduction}
\vspace{-3mm} In recent years, graph neural networks (GNNs) \cite{zhou2020graph} are shown to be effective in extracting information from graph structures like social networks with millions of nodes and billions of edges \cite{cui2018survey}. Indeed, GNNs account for over 90\% of the leading models in solving the open graph benchmark suite \cite{hu2020open,huang2021understanding}.

A GNN is designed to propagate and aggregate information across graph nodes. This is
achieved by applying a kernel function to a feature matrix of graph nodes, which captures the properties of nodes, as well as an adjacency matrix
that encodes the connectivity of graph edges. The kernel function is typically implemented using matrix multiplications \cite{zhou2020graph} that often dominate the GNN execution time during training and inference. Because most of the nodes in a real-life graph only have a small number of direct neighbors, the graph
adjacency matrix that a GNN kernel operates on is often sparse (i.e., many matrix elements are zeros). As a result, the matrix multiplication computation within a GNN is
essentially sparse matrix multiplication (SpMM) operations. 

There is an extensive body of work in optimizing SpMM for scientific workloads \cite{gilbert2008unified}. Various sparse matrix storage
formats have been proposed to reduce the memory and computation overhead of SpMM \cite{greathouse2014efficient,langr2015evaluation}.
Studies have also shown that choosing the right storage format can have a significant impact on the SpMM performance
\cite{mehrabi2021learning}. Although SpMM performance optimization is a well-studied field in traditional high-performance computing (HPC)
domains, the benefit of sparse matrix storage format selection is unclear on the new GNN workloads. Existing deep learning frameworks like
PyTorch \cite{paszke2019pytorch} and Tensorflow \cite{abadi2016tensorflow} all use a single, static sparse matrix storage format across
graph inputs. Since GNNs are becoming an important application class, it is essential to understand how GNN performance can benefit from
sparse matrix format selection.

This paper presents the first study of sparse matrix storage selection on GNN performance. We consider five representative GNN
architectures and six commonly used sparse matrix storage formats. We empirically
demonstrate that choosing a suitable sparse matrix storage format  can  have  a  significant  performance  benefit,  but  the  right  format
changes depending on the input matrix. We show that unlike traditional HPC workloads, the matrix sparsity can change over time as the GNN iterates over the input graph; and as a result, the suitable format can vary throughout GNN execution.

In light of this observation, we employ machine learning to automatically construct a predictive model based on XGBoost \cite{chen2015xgboost} for sparse matrix format selection.
Our predictor predicts, at runtime, the sparse matrix storage format and the associate SpMM computation kernel for each GNN kernel. Our predictor is first trained \emph{off-line} using synthetic matrix data. Then, using a set of automatically tuned features of
the matrix input, the predictor determines the optimal storage format to use before entering a kernel. We showcase that our approach is
generally applicable and can adapt to various optimization goals to find different trade-offs between the memory overhead and execution
time.

We evaluate our approach by applying it to five GNN architectures running on multi-core CPUs using both real-life and synthetic graph data.
We compare our approach against two prior machine-learning methods \cite{sedaghati2015automatic,pichel2019sparse} for selecting sparse
matrix storage formats. Experimental results show that our approach gives better performance over alternative optimization strategies by
giving an average 1.17x speedup. The performance of our approach translates to average 89\% of the oracle, a theoretically perfect
predictor for storage form selection (Section \ref{sec:oraclp}). performance given by a theoretically perfect predictor.

This paper makes
the following contributions:

\begin{itemize}
\item It is the first paper to study sparse matrix storage format selection on GNN performance;
\item It shows how machine learning techniques can be employed to develop a runtime predictor for optimizing GNN sparse matrix format selection;
\item It provides quantified performance results of widely used sparse matrix storage formats on representative GNN architectures.
\end{itemize}

\vspace{-3mm}
\section{Background}
\vspace{-2mm}
\subsection{Graph Neural Networks}
\vspace{-2mm}

A GNN operates on a graph structure, where each graph node is associated with a
$d$-dimensional feature vector of numerical values known as embeddings. Edges between nodes indicate their relationship, quantified with edge weights. 
For a graph with $N$ nodes, the graph edges
are encoded in an $N \times N$ adjacency matrix, $A$, and the node embeddings are stored in an $N \times d$ feature matrix, $X$. 

Like most neural networks, a GNN model can have multiple layers. Each layer is represented by two functions: i) an aggregation function
and ii) an update function (i.e., a combination function). During training, a GNN takes as input the adjacency matrix, $A$, of the graph.
It then uses a neighbourhood aggregation scheme to update the feature vector of each graph node based on the feature vector of its
neighboring nodes. Feature aggregation is performed by first applying the aggregation function (e.g., reductions)
to collect the features of the neighbours for a given node and then updating each node's feature vectors using the updating
function. After repeating this process of updating node features for a fixed number of times, a readout function is
applied to aggregate the feature matrix to a single numerical vector to be used as the graph representation.

The aggregation and update functions used by a GNN layer are implemented using matrix multiplications.
Because the graph adjacency matrix, $A$, is sparse in many real-life graphs, the GNN matrix multiplications are often realized as SpMM to reduce the memory footprint and processing time \cite{huang2021understanding}. When profiling 5 representative GNN models (Section \ref{sec:platform}) on real-life datasets, we find that SpMM can account for 95\% of the GNN processing time.

\vspace{-3mm}
\subsection{Sparse Matrix Storage Formats\label{sec:sf}}
\vspace{-2mm}
Our work considers the following commonly used sparse matrix storage formats:

\cparagraph{COO.} The coordinate list (COO) stores a list of (row, column, value) tuples of non-zero elements. This is the default storage format used by PyTorch-geometric \cite{fey2019fast} for graph processing. 

\cparagraph{CSR.} The compressed sparse row (CSR) format uses three arrays to represent non-zero matrix elements, that respectively contain non-zero values, the beginning position of each row, and the column indices of non-zero elements. CSR is similar to COO, but compresses the row indices, hence the name.

\cparagraph{CSC.} The compressed sparse column format (CSC) is similar to CSR, with one exception for using an array to store the target matrix's row indices of non-zero elements instead of column indices as in CSR.

\cparagraph{DIA.} The diagonal format (DIA) stores non-zero elements along the diagonal direction of a matrix into a row of a 2-dimensional array. It is best suited for non-zero elements that appear along the diagonals of a matrix.

\cparagraph{BSR.}
The block sparse row format (BSR) evenly divides the input matrix into blocks. It is CSR with dense sub-matrices of fixed shape instead of scalar items.

\cparagraph{DOK.}
The dictionary of keys format (DOK) stores key-value pairs $<$(row,column), value$>$ in a dictionary (e.g., a hash table). Elements that are not presented in the dictionary are treated as zero elements.

\cparagraph{LIL.}
The linked list (LIL) format stores non-zero elements and their column indices in a linked list. This format uses a row-based linked list, where each row is a list of column indices of non-zero elements.

\vspace{-3mm}
\section{Motivation} \label{moti}
\vspace{-2mm}
\begin{table}[t!]
\caption{Input matrix sparsity from graph datasets}
    \centering
    \scriptsize
    \begin{tabularx}{\textwidth}{XXXX}
    \toprule
    \textbf{Name} & \textbf{Adj. Matrix Density} & \textbf{Adj. Matrix Size} & \textbf{Node Feature Vector Dimension}\\
    \midrule
    \rowcolor{Gray} CoraFull & 0.6\% & $19,793\times8,710$  & 19,793\\
    Cora & 1.27\% & $2,708\times1,433$ & 2,708\\
    \rowcolor{Gray} DblpFull  & 0.31\% & $17,716\times1,639$ & 17,716 \\
    PubmedFull & 10.02\% & $19,717\times500$ & 19,717\\
    \rowcolor{Gray} KarateClub & 2.94\% & $34\times34$ & 34\\
    \bottomrule
    \end{tabularx}
    \label{tab:Dataset_Detail}
    \vspace{-1mm}
\end{table}

\begin{figure}[t!]

\begin{minipage}[t]{0.5\linewidth}
\includegraphics[width=\textwidth]{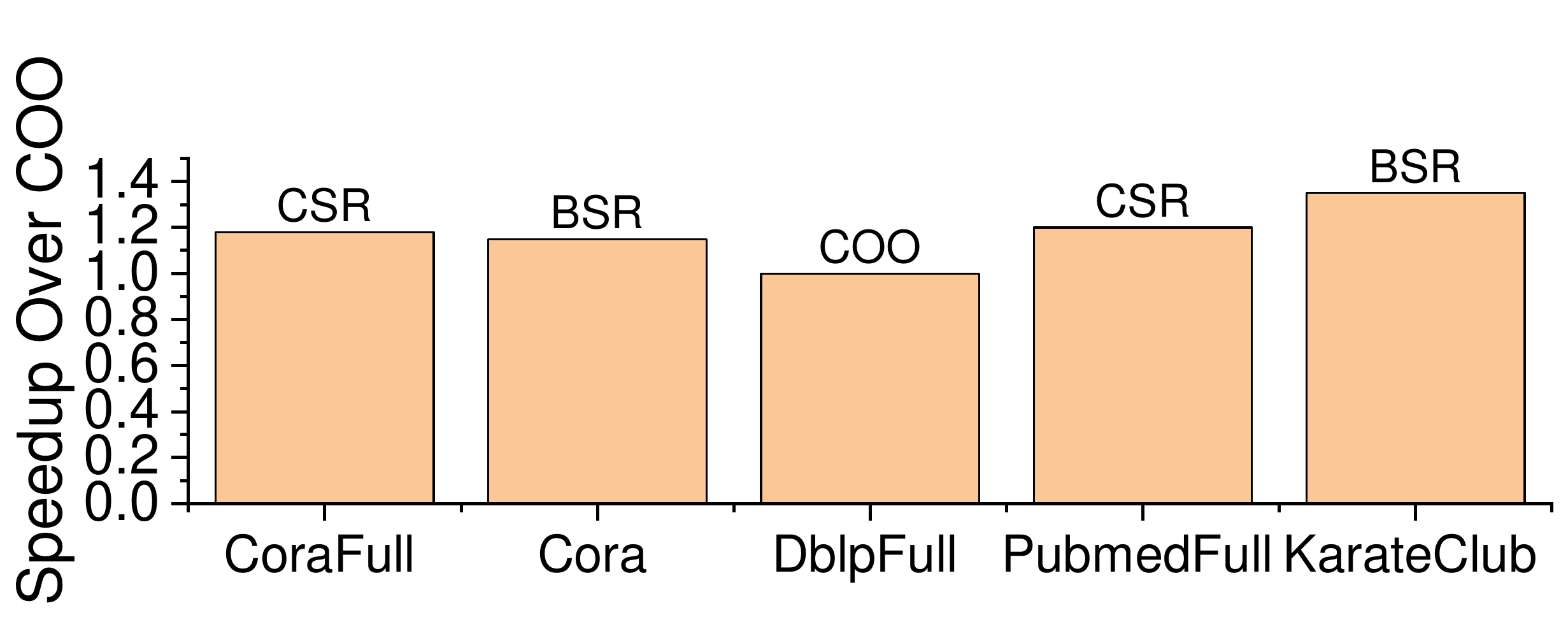}
\caption{The best-performing storage format per dataset.}
\label{fig:beststoring_withdiffdataset}
\vspace{-1mm}
\end{minipage}
\begin{minipage}[t]{0.5\linewidth}
\centering
\includegraphics[width=\textwidth]{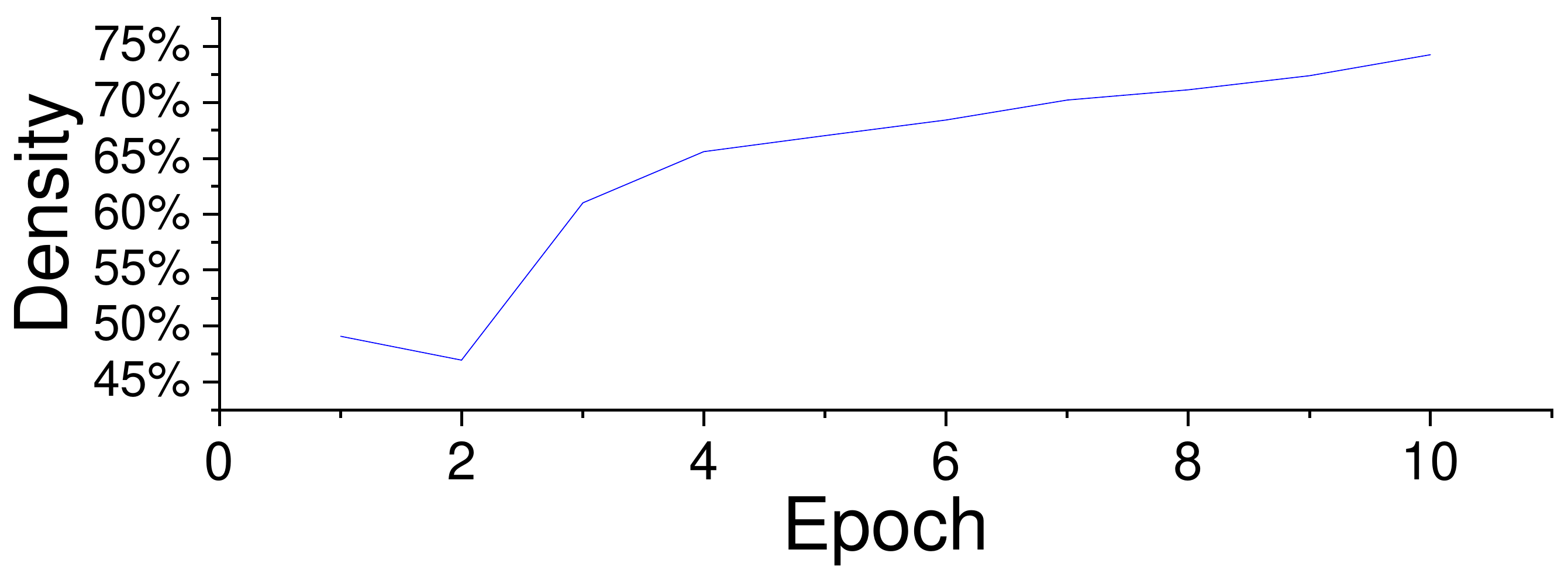}
\caption{Changes of the adjacency matrix density over GNN training epochs. }
\label{fig:LayerSparsity}
\end{minipage}
\vspace{-4mm}
\end{figure}

\vspace{-2mm}
As a motivating example, consider applying a
two-layered  graph convolution network (GCN) model ~\cite{kipf2016semi} to 5 real-life graph datasets (Table \ref{tab:Dataset_Detail}) using the 7 sparse matrix storage formats described in Section \ref{sec:sf}.

\vspace{-4mm}
\subsection{Setup}
\vspace{-2mm}
In this experiment, we consider five real-life graph datasets used in prior work
\cite{bojchevski2017deep}. Table \ref{tab:Dataset_Detail} summarizes the size and sparsity of the graph adjacency matrix, and the
dimension of the node feature vector (a dense vector). We
run the GCN model on a 2.0 GHz 20-core Intel Xeon CPU. We note that it is common to run a GNN on the CPU due to the large
memory footprint of graph processing \cite{bojchevski2017deep}.

\vspace{-4mm}
\subsection{Results}
\vspace{-2mm}
Figure \ref{fig:beststoring_withdiffdataset} shows the best-performing sparse matrix format for each dataset, when a format is used to encode the initial model input and used throughout the model training process. Here, we normalize the measured runtime against the time of the PyTorch-geometric default COO format. While COO gives the best
performance on \texttt{DBLPFull}, it leaves much room for performance improvement on other datasets. Furthermore, we also observe that the best-performing storage format varies depending on the input
dataset.

\begin{figure}[t!]
\centering
\subfigure[CoraFull] {\label{fig:CoraFull_b}
    \begin{minipage}[t]{0.5\linewidth}
        \centering
        \includegraphics[width=1\textwidth]{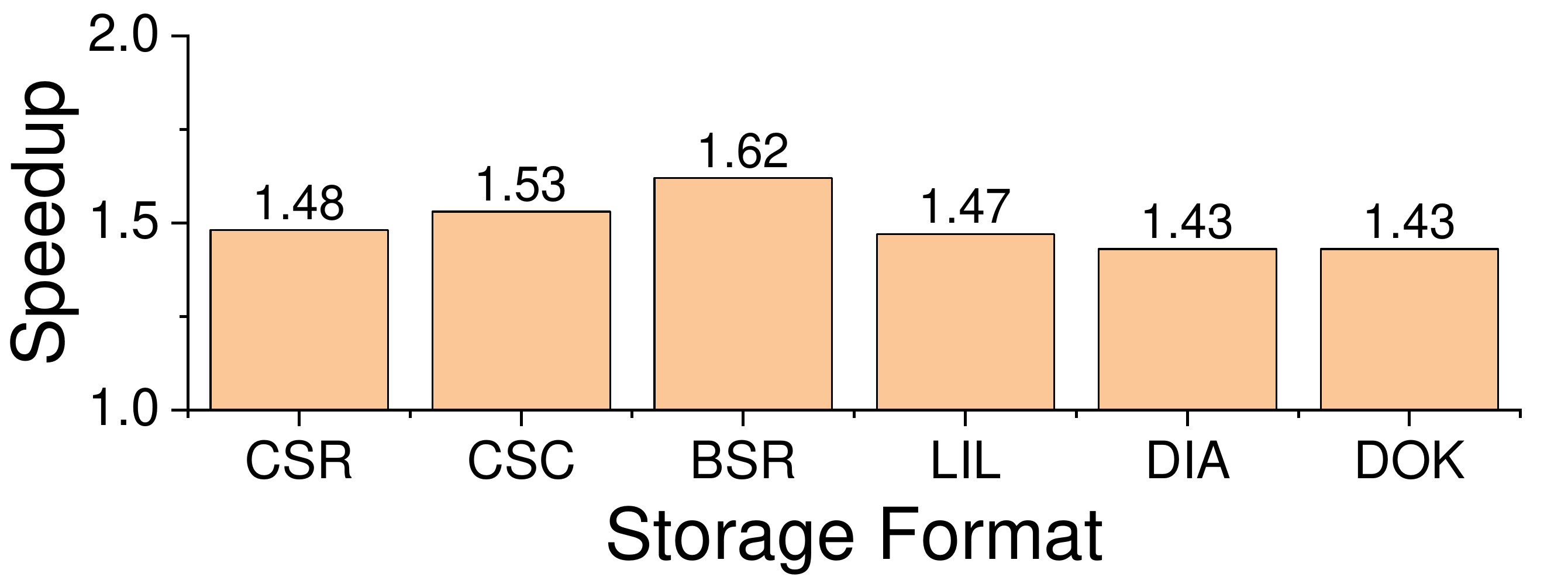}\\
    \end{minipage}%
}%
\subfigure[PubmedFull] { \label{fig:pubmedfull_b}
    \begin{minipage}[t]{0.5\linewidth}
        \centering
        \includegraphics[width=1\textwidth]{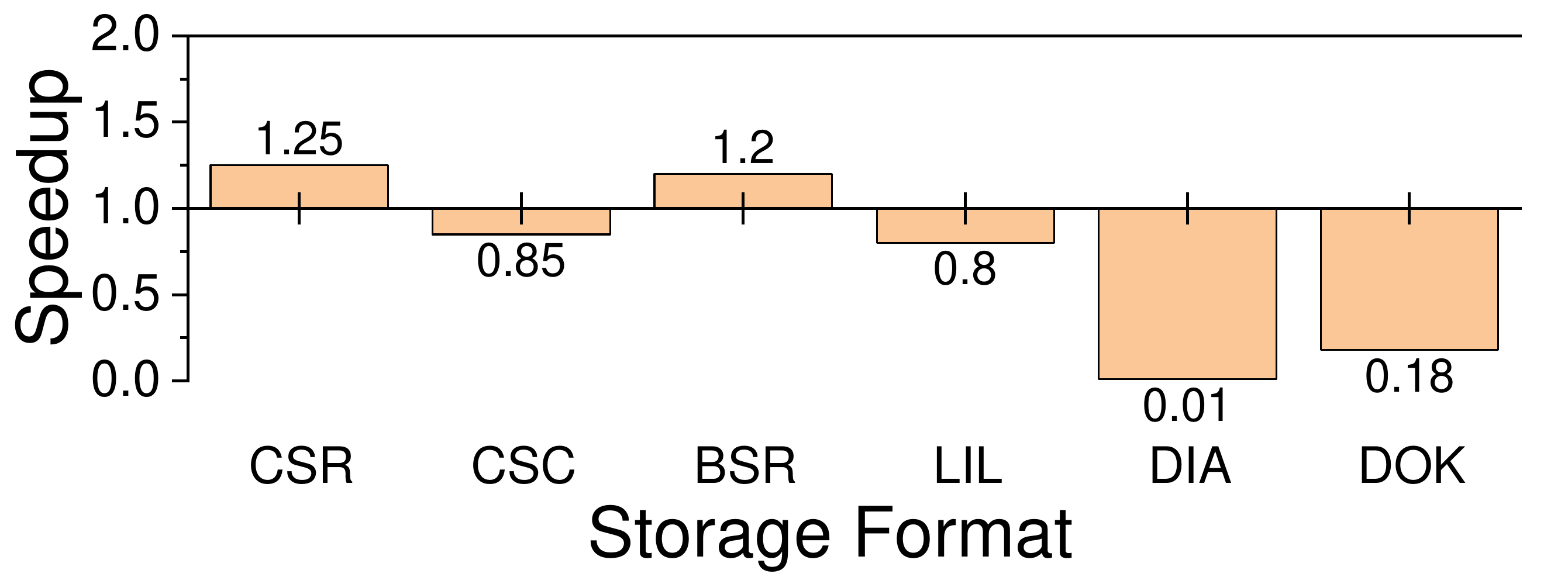}\\
        \vspace{0.05cm}
    \end{minipage}%
}%
\vspace{-3mm}
\caption{Performance improvement over the PyTorch-geometric default COO format on the CoraFull (a) and PubmedFull dataset (b) when using different sparse matrix format
to store the output of the first GNN layer.}
\label{fig:CoraFull_pubmedfull_b}
\vspace{-4mm}
\end{figure}

If we now consider Figure \ref{fig:LayerSparsity}, we see that the density of the input matrix increases as we iterate over the GNN model
on the \texttt{CoraFull} dataset. This is expected as a GNN tries to incorporate further neighbourhood information by iterating over the
graph, which in turn increases the reach and information propagation of a graph node. As can be seen in figure \ref{fig:CoraFull_pubmedfull_b}, CSR is the best format used to store the
neural network input (i.e., the feature and the adjacency matrix) for both the \texttt{CoraFull} and \texttt{PubmedFull} datasets. Thus, for a
model with a single layer GNN, CSR might be the best storage format. However, for a typical GNN model with multiple GNN layers, the
sparsity of the matrices processed by the latter layers can change, calling for a different storage format to be used. Specifically, for
\texttt{CoraFull} (figure \ref{fig:CoraFull_b}) used in our setting, using CSC, LIL and DIA after the first GNN layer can also give a
relatively good speedup over COO, but these format give no benefit on \texttt{PubmedFull} (Figure \ref{fig:pubmedfull_b}) because of the
changing distribution of the non-zero elements, the details can be seen in figure \ref{fig:CoraFull_pubmedfull_b}.

\cparagraph{Lesson learned.} This example shows that choosing the right sparse matrix storage format can have a significant performance
benefit, but the choice depends on the input data and the GNN layers. Therefore, the decision for storage format should be made on a per GNN layer basis during runtime.

\vspace{-4mm}
\section{Our Approach}
\vspace{-4mm} Our work aims to choose the most efficient sparse matrix storage format for accelerating GNN performance or finding a
trade-off between the memory footprint and runtime. As the right choice depends on the characteristics of the input matrix processed by a
GNN layer, and the optimal storage format can change over the duration of the training, we wish to develop an approach to automatically
derive a storage format (and the SpMM kernel) on a per input basis.

To this end, we employ machine learning to build a classifier to predict the sparse matrix storage format to use from a pool of candidate
formats. The predictive model takes as input a feature vector of numerical values, which describe the essential characteristics of the
input matrix. It then produces a label, indicating which of the storage formats to be used by a GNN layer. We provide APIs (Section \ref{sec:deployment}) to monitor the input matrix sparsity and dynamically adjust the storage format to use before entering a GNN layer at runtime. If the chosen format is
different from the one used by the previous layer or a prior training epoch, our library will convert the input matrix to the chosen format. Note that we include the overhead of format conversion and feature extraction in all our experimental results.

\vspace{-2mm}
\subsection{Predictive Modeling}
\vspace{-2mm}
Our predictive model builds upon the XGBoost classifier \cite{chen2015xgboost}. We have evaluated a number of alternative classification
techniques, including multilayer perceptron (MLP) neural networks, K-Nearest neighbour (KNN), and support vector machines (SVM).  We
choose XGBoost because of its good generalization ability \cite{chen2015xgboost}, its decision-tree-like structure is interpretable, and
its better and more robust performance over alternatives on our problem (Section \ref{sec:optp}). In the remainder of this section, we describe our predictive model by following the classical 4-step process for supervised
learning: i) problem modeling, ii) training  data generation, iii) train a predictor and iv) implement the predictor.

\vspace{-2mm}
\subsection{Problem Modeling}
\vspace{-2mm}
\label{sec:ms}
\begin{figure}[t!]
\begin{minipage}[t]{0.5\linewidth}
\centering
\includegraphics[width=\textwidth]{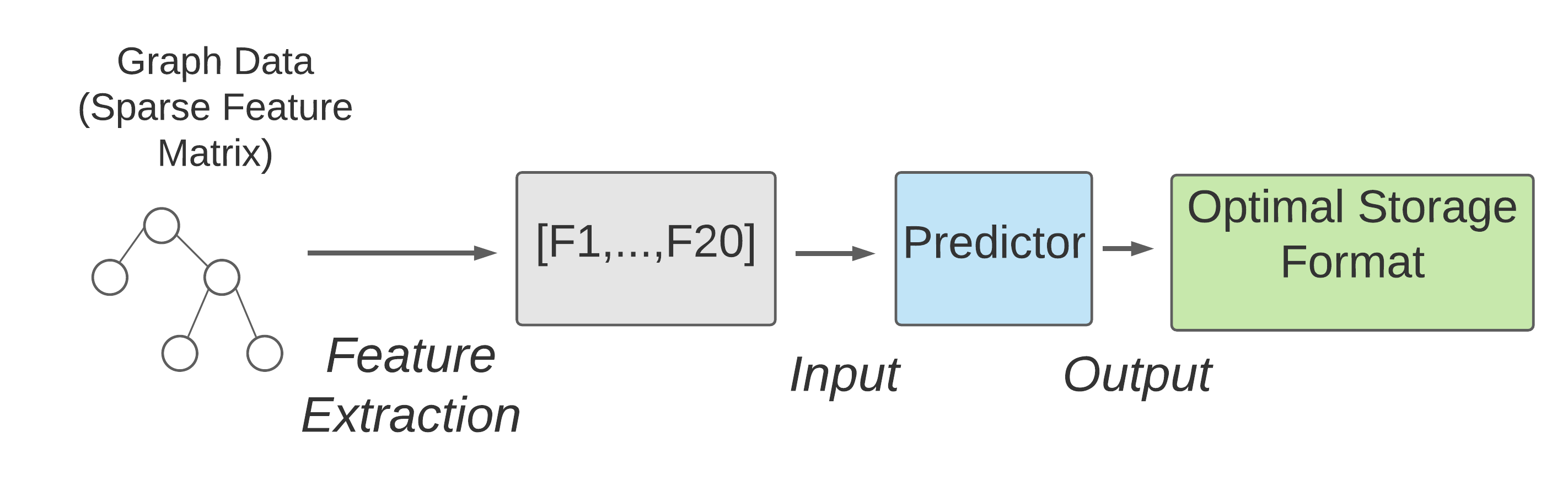}
\vspace{-4mm}
\caption{Overview of our predictive model for choosing sparse matrix storage format.}
\label{fig:approach_overview}
\end{minipage}
\begin{minipage}[t]{0.5\linewidth}
\centering
\includegraphics[width=\textwidth]{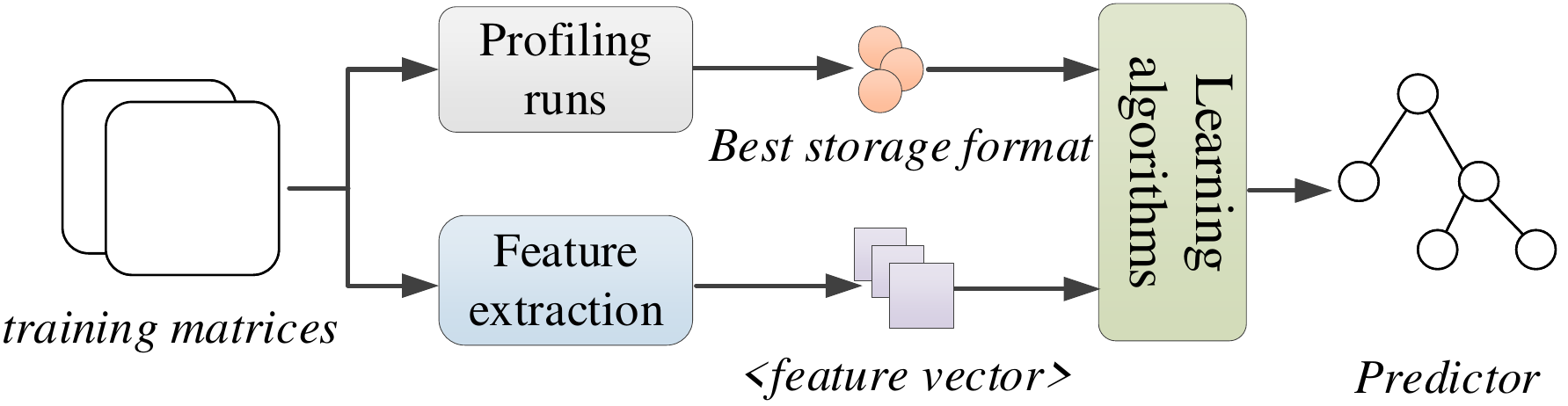}
\vspace{-4mm}
\caption{Overview of our training process}
\label{fig:training}
\end{minipage}
\vspace{-3mm}
\end{figure}

Figure \ref {fig:approach_overview} depicts the workflow of our approach. The deployed model extracts features from the adjacency
and feature matrices and uses the feature values to predict the sparse matrix storage format to use. Our library automatically converts the input matrix to the selected storage format if needed.
Note that a SpMM computation kernel can be chosen based on the object type of the input. Since we implemented our prototype in PyTorch, this computation kernel selection process is performed automatically by the Python library.

As depicted in Figure \ref{fig:training}, our model is trained offline using training samples. The trained model can be applied to
any previously unseen matrix. Training involves finding the best storage format, extracting feature values for each training
matrix and learning a model from the training data, described as follows.

\vspace{-2mm}
\subsection{Training Data Generation}\label{sec:dg}
\vspace{-1.5mm}

We use 300 synthetically generated square matrices to train the XGBoost model. The matrix size of our training samples ranges from $1,000$
to $15,000$, increased with a step of 200. We populate the matrix with random values of 0 and 1 with a sparsity ranging from 0.1\% to 70\%,
to simulate the matrix sparsity seen at the initial model graph input and later message propagation stages.
For each training matrix, we exhaustively execute the
SpMM computation kernel with each sparse matrix storage format and record the best performing format for each matrix sample on each kernel. We then label each best-performing configuration with a unique number (i.e., class label). Note that we apply cross-validation in our evaluation to make sure we always test the trained model on unseen datasets.

\cparagraph{Optimization goal.} Our approach allows the user to find a trade-off between the memory footprint and the GNN performance and
train a predictive model for their optimization goal. Specifically, in this work, we consider the following optimization formulation,
but other formulas can also be used:
\begin{equation}
\small
\min_{O} O_{l \in L} = w \times R + (1.0 - w) \times M
\label{equ:nomarlization}
\end{equation}
where $R$ and $M$ are the normalized running time and memory footprint for a sparse matrix storage format from a collection of candidate
formats ($L$), and $w$ is a configurable weight parameter. Note that we scale the execution time and memory footprint to the $(0,1)$ range
using the min-max values found from the profiled training data. Essentially, our goal is to minimize the weighted sum, $O$ in
Eq \ref{equ:nomarlization} to trade runtime for a lower memory footprint. For example, setting $w$ to $0$ and $1.0$ means we only optimize for memory overhead and speeds respectively.

\begin{figure}[t!]
\centering
\includegraphics[width=0.75\textwidth]{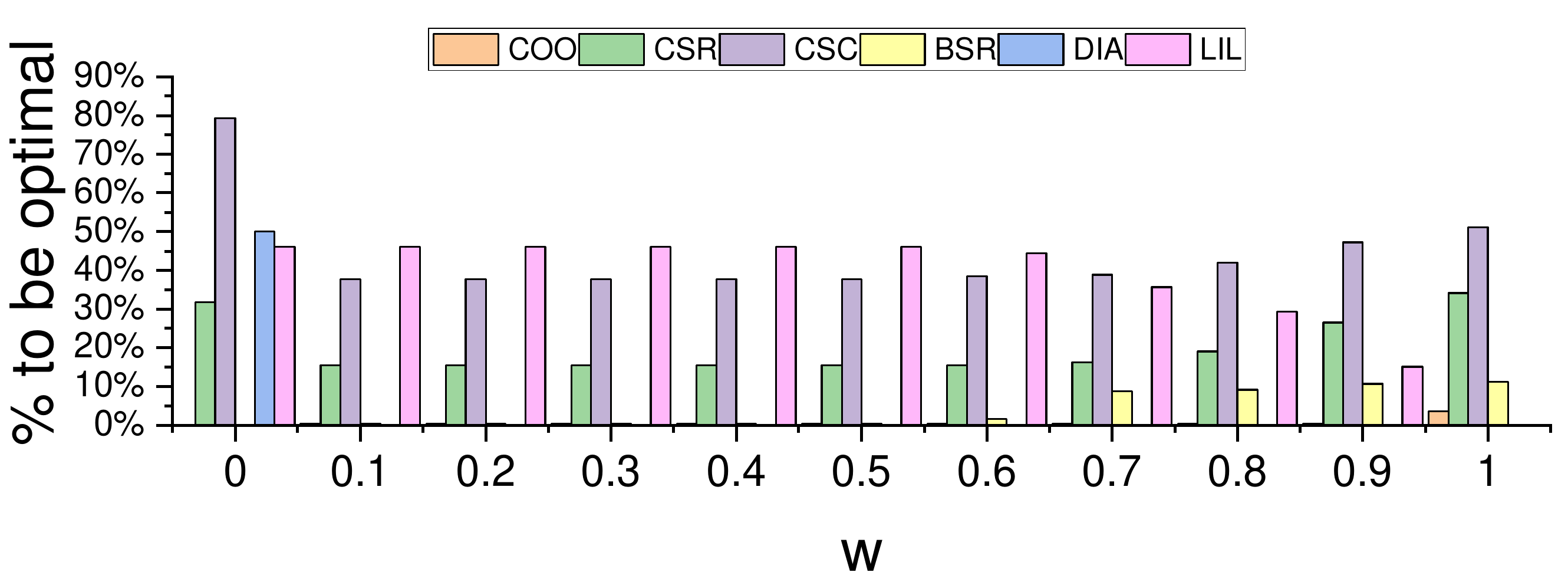}
\vspace{-5mm}
\caption{How often a storage format is considered to be optimal on our synthetic training data when varying the weight $w$ in Eq \ref{equ:nomarlization}. Noted that there might be multiple optimal formats for a single input if the final output $O$ is very similar ($\pm 0.0001$). }
\label{fig:frequency}
\vspace{-4mm}
\end{figure}

Our training data includes the raw measurements of the execution time and memory footprint for each storage format under each matrix. We then apply the Eq \ref{equ:nomarlization} to label the storage format that gives the smallest $O$ for each training
sample. Figure \ref{fig:frequency} lists the frequency of a storage format to be found to be optimal on our training dataset. Here, the x-axis shows different settings of $w$ in Eq \ref{equ:nomarlization}. As can be seen from the diagram, the optimal storage format can vary depending on the optimization criterion. Our approach can adapt to such changes by automatically learning from the training samples (see Section \ref{sec:tm}).

 For each training data
sample, we also extract the values of a selected set of features (described in Section \ref {sec:fe}). We note that training is a one-off cost, and the trained predictive model can be used by any GNN model to optimize the SpMM computation kernel.

\vspace{-2mm}
\subsection{Feature Engineering\label{sec:fe}}
\vspace{-2mm}

\begin{table}[t!]
\caption{Matrix feature used by in our predictive model}
    \centering
    \scriptsize
    \begin{tabularx}{\textwidth}{ssb||ssb}
    \toprule
    \textbf {No.}& \textbf{Featur.} & \textbf{Description} &  \textbf {No.}& \textbf{Featur.} & \textbf{Description} \\
    \midrule
    \rowcolor{Gray} F1 & numRow & \# rows &
    F2 &  numCol & \# columns \\
    F3 & NNZ & \# Non-zeros &
    F4 & N\_diags & \# diagonals \\
	 \rowcolor{Gray} F5 & aver\_RD &  Avg. \# non-zero elements per row &
    F6 & max\_RD &  Max. \# non-zeros per row  \\
     F7 & min\_RD &  Min. \# non-zeros  per row  &
     F8 & dev\_RD &  Standard deviation of non-zero numbers per row  \\
   \rowcolor{Gray} F9 & aver\_CD &  Avg. \# non-zeros  per column  &
    F10 & max\_CD &  Max.  \# non-zero values per column  \\
      F11 & min\_CD &  Min. \# non-zero values per column &
    F12 & dev\_CD &  The deviation number of non-zeros per column\\
    \rowcolor{Gray} F13 & ER\_DIA &  Ratio of non-zeros in diagonals  &
    F14 & ER\_CD &  Ratio of non-zeros in column-packed structure \\
    F15 & row\_bounce & Avg. differences between non-zeros of adjacent rows &
    F16 & col\_bounce &  Avg. difference between non-zeros of adjacent columns \\
    \rowcolor{Gray} F17 & density &  Density of non-zeros &
    F18 & cv &  Normalized variation of non-zeros per row \\
    F19 & max\_mu &  max. RD - avg. RD \\
    \bottomrule
    \end{tabularx}
    \label{tab:Feature_Detail}
    \vspace{-6mm}
\end{table}


\cparagraph{Feature selection.}
A key aspect in building a good machine learning predictor is finding the right representation, or \emph{features}, to capture
the essential characteristics of the input workload.
 We start by considering over 30 raw features chosen based on previous work of SPMV optimization \cite{sedaghati2015automatic}. Most of the features are used to capture the distribution of non-zero elements of the input matrix, which can be extracted
in parallel to reduce the overhead of feature extraction.

To learn effectively over a small training dataset, we use the feature score given as a by-product of the XGBoost training process to  select a compact set of features. The feature score is computed summing up how many times each feature is split on the decision tree. We then keep features that contribute to 95\% of the aggregated importance scores across all raw features.
Using a fewer number of features also help us to reduce the overhead of
runtime feature extraction. Table \ref{tab:Feature_Detail} summarizes our chosen matrix features.

\cparagraph{Feature normalization.}
In the final step, we scale each of the extracted feature values to a common range (between 0 and 1) to prevent the range of any single
feature from being a factor in its importance. We record the minimum and maximum values of each feature in the training dataset in order to
scale the feature values of an unseen matrix. We also clip a feature value to make sure it is within the expected range during deployment.

\begin{figure}[t!]
\centering
\includegraphics[width=0.75\textwidth]{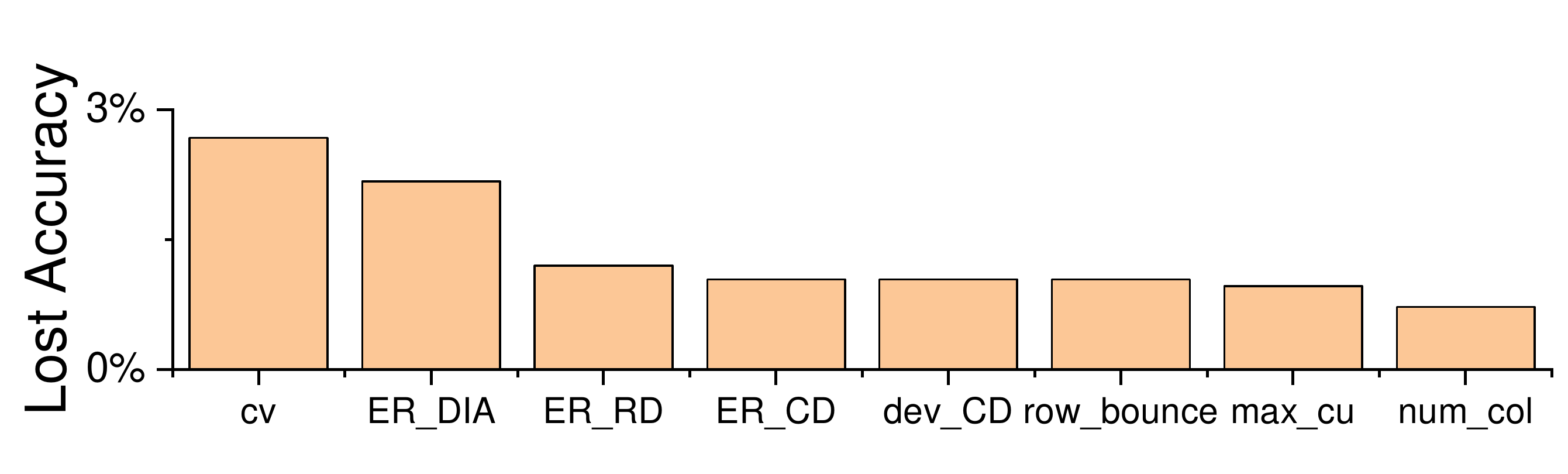}
\vspace{-4mm}
\caption{Top-8 features which can lead to a high loss
in accuracy if they are not used.}
\vspace{-5mm}
\label{fig:acclossforfeature}
\end{figure}

\cparagraph{Feature importance.}
Figure \ref{fig:acclossforfeature} shows the top 8 dominant features based on their impact on our predictive model accuracy. We calculate
feature importance by first training a model using all 19 of our chosen features, and record the accuracy of our model. In turn, we then
remove each of our features, retraining and evaluating our model on the other 18, noting the drop in prediction accuracy. We then normalize
the values to produce a percentage of importance for each of our features. Features for measuring the non-zero element distribution, like
ER\_DIA and cv in Table \ref {tab:Feature_Detail}, are important for choosing the storage format. The similar distribution of feature importance is
an indication that each of our features is able to represent distinct information about the matrix workload, all of which is important for
the prediction task at hand.

\vspace{-3mm}
\subsection{Training The Model} \label{sec:tm}
\vspace{-2mm} The collected feature values, together with the desired label for each training matrix, are passed to a supervised learning
algorithm to learn the XGBoost model. The time for training the predictor is dominated by generating the training data. In this work, it
takes less than a week to label all the training samples using a single multi-core server.  In comparison, processing the raw data and
building the models took a negligible amount of time, less than an hour run in a RTX 2060 GPU. Since training is only performed once, it is
a \emph{one-off} cost.

\vspace{-3mm}
\subsection{Using The Model\label{sec:deployment}}
\vspace{-2mm} The trained predictor can be applied to a new, unseen matrix used by a SpMM kernel. We implement our predictive model using
the Python Scikit-learn \cite{sklearn_api} package, which can be easily integrated with mainstream deep learning frameworks. We have
encapsulated all of the inner workings, such as feature extraction, prediction and storage format conversion and kernel selection, into a
single package. Prediction is done by calling a dedicated \texttt{SpMMPredict} function (provided by our library) before each GNN layer.
The function takes as input a matrix object and outputs a matrix object stored using the predicted storage format. Depending on the matrix
object type, the corresponding SpMM kernel will be automatically chosen. Our current implementation supports PyTorch, but it can be easily
ported to other deep learning frameworks.

\vspace{-2mm}
\section{Experimental Setup}
\vspace{-2mm}
\subsection{Software and Hardware} \label{sec:platform}
\vspace{-3mm}
 \cparagraph{Evaluation platform.} Our hardware platform is a dual-socket multi-core server with two 20-core Intel Sky Lake Xeon Gold 6138
CPUs running at 2.0 Ghz with 192GB of RAM.
 Our evaluation platform runs Centos 7 with Linux kernel version 3.10. We test our approach on PyTorch
v1.4.0, running on the CPU.

\cparagraph{GNN models.}We apply our approach to 5 representative GNN architectures, including GCN, graph attention network (GAT)
\cite{gat2018graph}, relational graph convolutional neural network (RGCN) \cite{schlichtkrull2018modeling}, GNN with feature-wise linear
modulation (FiLM) \cite{brockschmidt2020gnn} and efficient graph convolutions (EGN)\cite{tailor2021adaptive}. We use the open-source
implementation provided by PyTorch-geometric library \cite{fey2019fast} by stacking two GNN layers to form a standard graph model.

\cparagraph{Datasets.} In our evaluation, we use two graph data suites, CoraFull \cite{xu2019crosslingual} and Entities \cite{schlichtkrull2018modeling}, containing a
total of 5 graph datasets with matrix sizes ranging from 19,793 to 58,086. To evaluate the generalization ability of our approach, we also
apply our approach to 100 synthetic matrices of different sizes and sparsity. For the synthetic data, we initialize weights in the adjacency matrices
by populating them with random single floating numbers between 0 and 1.0.

\vspace{-3mm}
\subsection{Evaluation Methodology\label{sec:priorwork}}
\vspace{-3mm} \cparagraph{Competitive methods.} We compare our approach against two closely related predictive methods for using machine
learning to choose the sparse matrix storage format. The first approach employs a convolutional neural network (CNN)
\cite{zhao2018bridging,pichel2019sparse}, and the second uses a decision tree model for format selection \cite{sedaghati2015automatic}. We
use an open-source implementation of ResNet \cite{paszke2019pytorch} as the CNN model.  To provide a fair comparison, we train all machine
learning models on the same training dataset using the methodology described in the source publications.

\cparagraph{Performance report.} We consider the end-to-end execution time, including the overhead of our predictive model (i.e., the time
spending on feature extraction, storage format transformation and model prediction). Our feature extraction process runs in parallel using
all CPU cores.   We measure the end-to-end training time by training each model on each dataset for 10 epochs. We run each matrix input 5
times and report the \emph{geometric mean} of the end-to-end training time and show the variations across different runs as a min-max bar.
Note that we only need to decide the matrix storage format once for each GNN layer across training epochs. Given that in our
evaluation, the sparse matrix distribution is similar across training epochs, and hence the overhead of our approach can be further
amortised across multiple training epochs.

\vspace{-4mm}
\section{Experimental Results}
\vspace{-3mm}
\subsection{Overall Results}
\vspace{-2mm}

\begin{figure}[t!]
\centering
\subfigure[Per GNN model] {
    \begin{minipage}[t]{0.5\linewidth}
        \centering
        \includegraphics[width=1\textwidth]{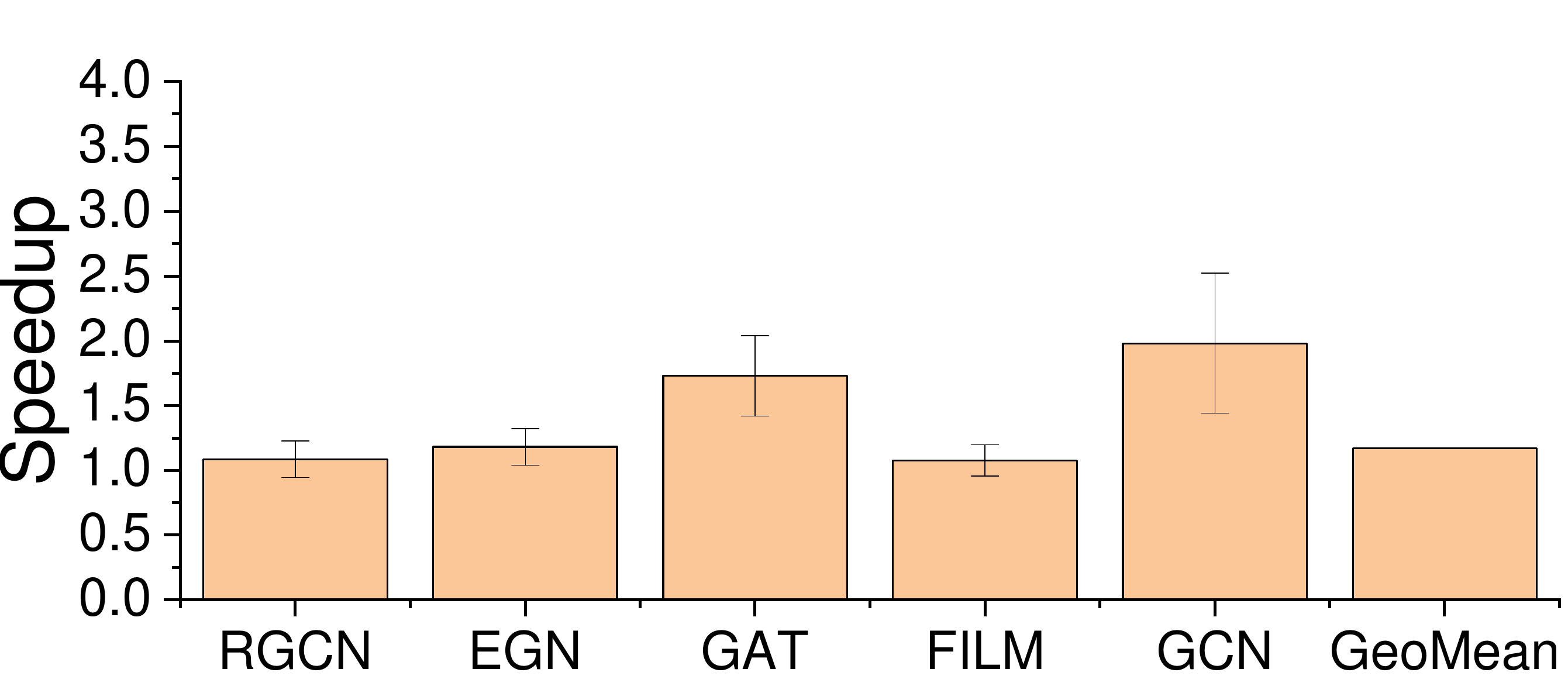}\\
        \label{fig:min_max_bar}
    \end{minipage}%
}%
\subfigure[Per Dataset] {
    \begin{minipage}[t]{0.5\linewidth}
        \centering
        \includegraphics[width=1\textwidth]{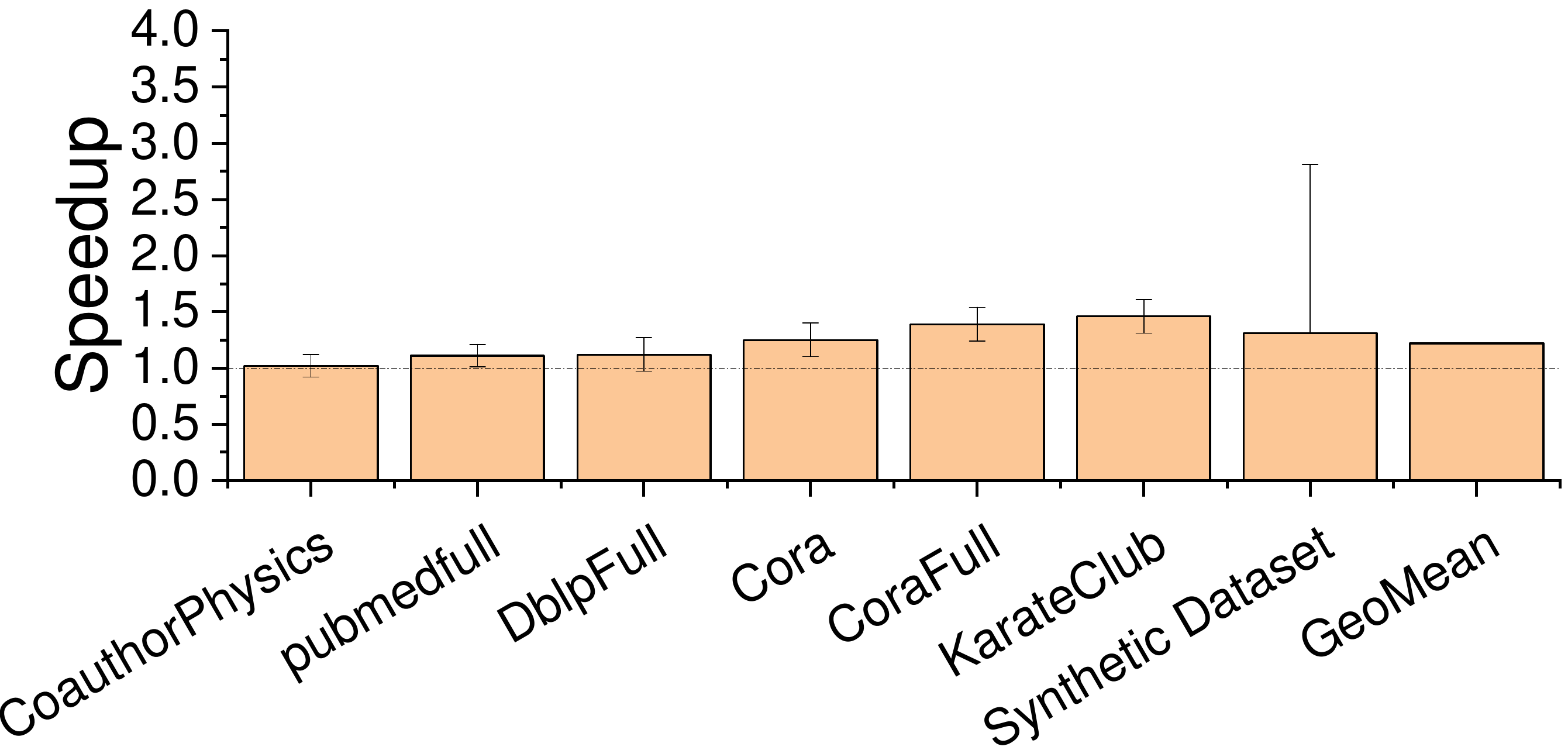}\\
        \vspace{0.05cm}
        \label{fig:evaluation}
    \end{minipage}%
}%
\vspace{-3mm}
\caption{Speedup given by our approach over COO. GeoMean represents the geometric mean given by the previous performance. }
\vspace{-3mm}
\label{fig:MinMaxBar}
\end{figure}

Figure \ref{fig:min_max_bar} shows the speedup over the PyTorch COO sparse matrix storage format for each GNN model across our evaluation datasets. Here, the min-max bar show the variance across the evaluated datasets. In this experiment, we aim to optimize for speedups by setting $w$ of Eq. \ref{equ:nomarlization}. Moreover, in Section \ref{sec:optp} we show our approach can generalize to other settings of $w$.

As can be seen from the diagrams, choosing the right sparse matrix storage format can improve the GNN performance. Our approach delivers an average speedup of 1.3x (up to 3x) on GCN, which involves many SpMM computations when performing the graph convolution operations. Our approach gives less performance improvement on RGCN because the dataset that RGCN operates is a dense edge-based dataset that does not benefit from sparse matrix format selection. Furthermore, on a small number of datasets, where the COO is the best format, our approach shows a minor slowdown, less than 7\%, due to the overhead of feature extraction. But for the majority of the evaluated datasets, our approach gives a noticeable improvement over COO. Overall, our techniques give an average speedup of 1.17x across GNN models and evaluation datasets.

Figure \ref{fig:evaluation} shows the achieved performance per real-world graph dataset across models. For most of the datasets, our approach gives noticeable speedups across GNN.

\vspace{-2mm}
\subsection{Compare to Prior Methods}
\vspace{-2mm}
Table \ref{tab:sota_compare} compares our approach against a CNN and a decision tree model for choosing the matrix storage format, where our approach gives a better overall prediction accuracy. The CNN model gives a poor prediction accuracy when the model is trained on 300 synthetic matrices. While the performance of the CNN model can be improved by using more training data, doing so would incur a higher overhead. Table \ref{tab:sota_compare} confirms that a higher prediction accuracy does translate into better speedup performance, where our approach improves the CNN and the decision tree model by 27\% and 3\%, respectively.

\begin{table}[t!]
\caption{Comparing our XGBoost approach with prior work}
\vspace{-2mm}
    \centering
    \scriptsize
    \begin{tabular}{lrrr}
    \toprule
    \textbf{Model} & \textbf{Inference Time (s)} & \textbf{Prediction Accuracy (\%)} & \textbf{Realized Speedup}\\
    \midrule
    XGboost (ours) & 0.0008 & 89.1 & 1.17\\
    CNN \cite{zhao2018bridging,pichel2019sparse} & 0.002 & 66.8 & 0.86\\
    Decision-Tree \cite{sedaghati2015automatic} & 0.0002 & 83.8 & 1.14\\
    \bottomrule
    \end{tabular}
    \label{tab:sota_compare}
    \vspace{-3mm}
\end{table}

\vspace{-2mm}
\subsection{Compare to Oracle Performance\label{sec:oraclp}}
\vspace{-2mm}
\begin{figure*}[t!]

\begin{minipage}[t]{0.5\linewidth}
\includegraphics[width=\textwidth]{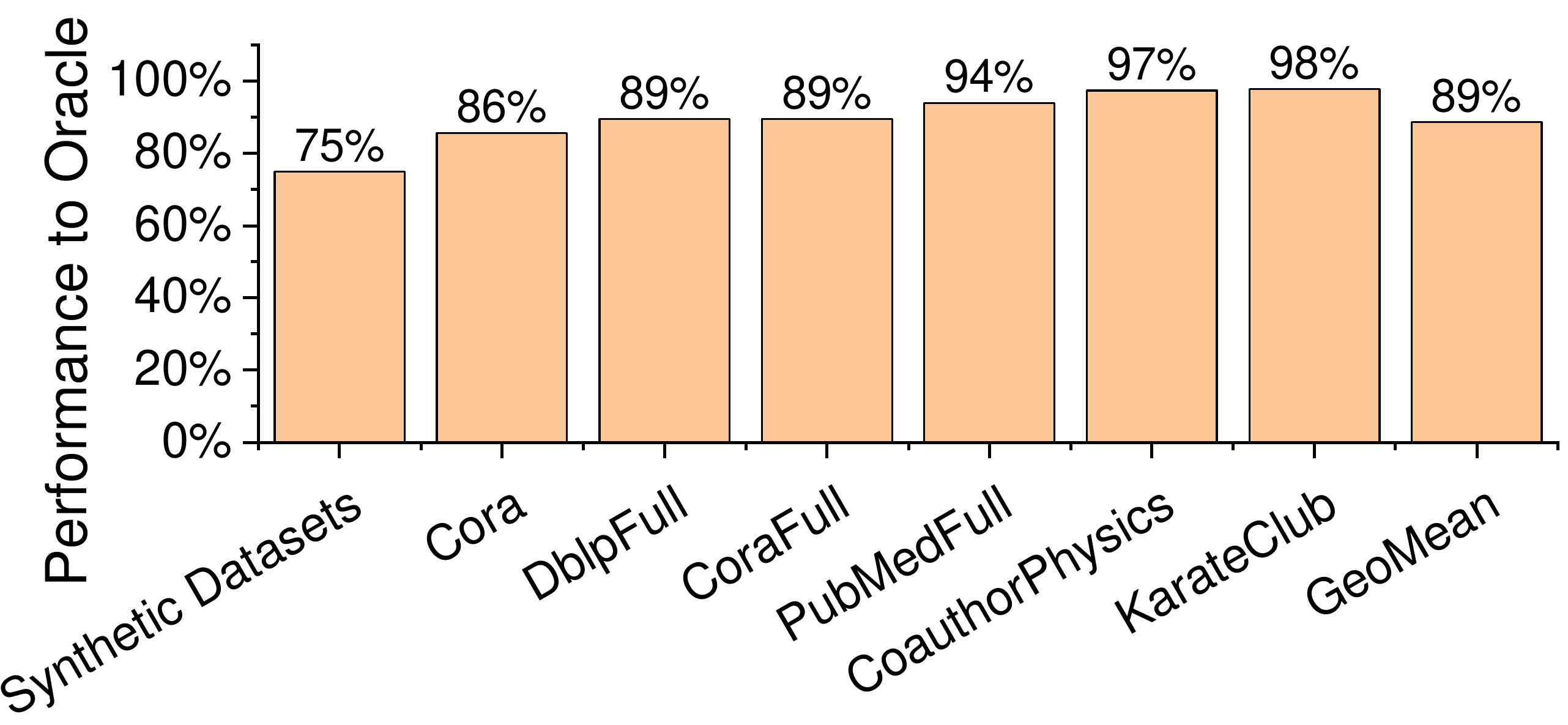}
\vspace{-5mm} \caption{Performance of our approach related to the Oracle performance.} \label{fig:oracle}
\end{minipage}
\begin{minipage}[t]{0.5\linewidth}
\includegraphics[width=1\textwidth]{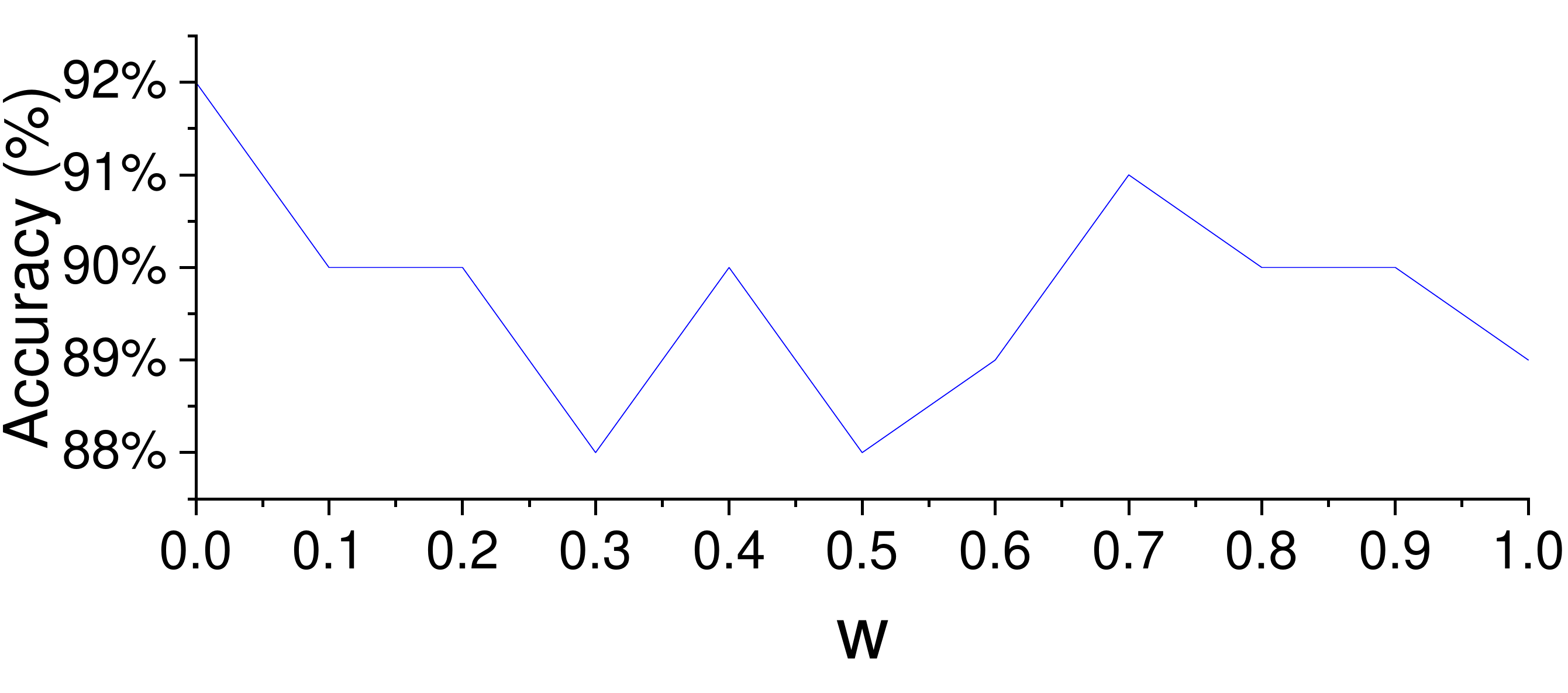}
\vspace{-6mm} \caption{Prediction accuracy of our approach when varying $w$ in Eq \ref{equ:nomarlization}. } \label{fig:normalized_acc}
\vspace{-9mm}
\end{minipage}
\vspace{-3mm}
\end{figure*}

Figure \ref{fig:oracle} compares our approach against \emph{a theoretically perfect predictor} for storage form selection, for which we call \emph{oracle}. We obtain the oracle performance by exhaustively profiling all candidate storage formats for each GNN layer to find out the best-performing format. The results show how close our predictive modeling approach is to the theoretical upper bound. Our approach achieves, on average,  89\% of the oracle performance. Our model can be further improved by using more training samples together with more representative features to characterise some of the input matrices better to improve the prediction accuracy.

\vspace{-3mm}
\subsection{Model Analysis\label{sec:optp}}
\vspace{-3mm} \cparagraph{Impact of optimization goal.} Our evaluation so far set $w$ to 1 of our optimization function
(Eq.~\ref{equ:nomarlization}) by solely optimizing for speeds. Figure \ref{fig:normalized_acc} shows prediction accuracy when we vary the
parameter settings. Our approach has a good generalization by giving the average accuracy of 90\%. This experiment shows that our approach
is flexible and can adapt to different optimization trade-offs.

\begin{figure}[t!]
\centering
\subfigure[Prediction Accuracy] {\label{fig:predictionAcc}
    \begin{minipage}[t]{0.5\linewidth}
        \centering
        \includegraphics[width=1\textwidth]{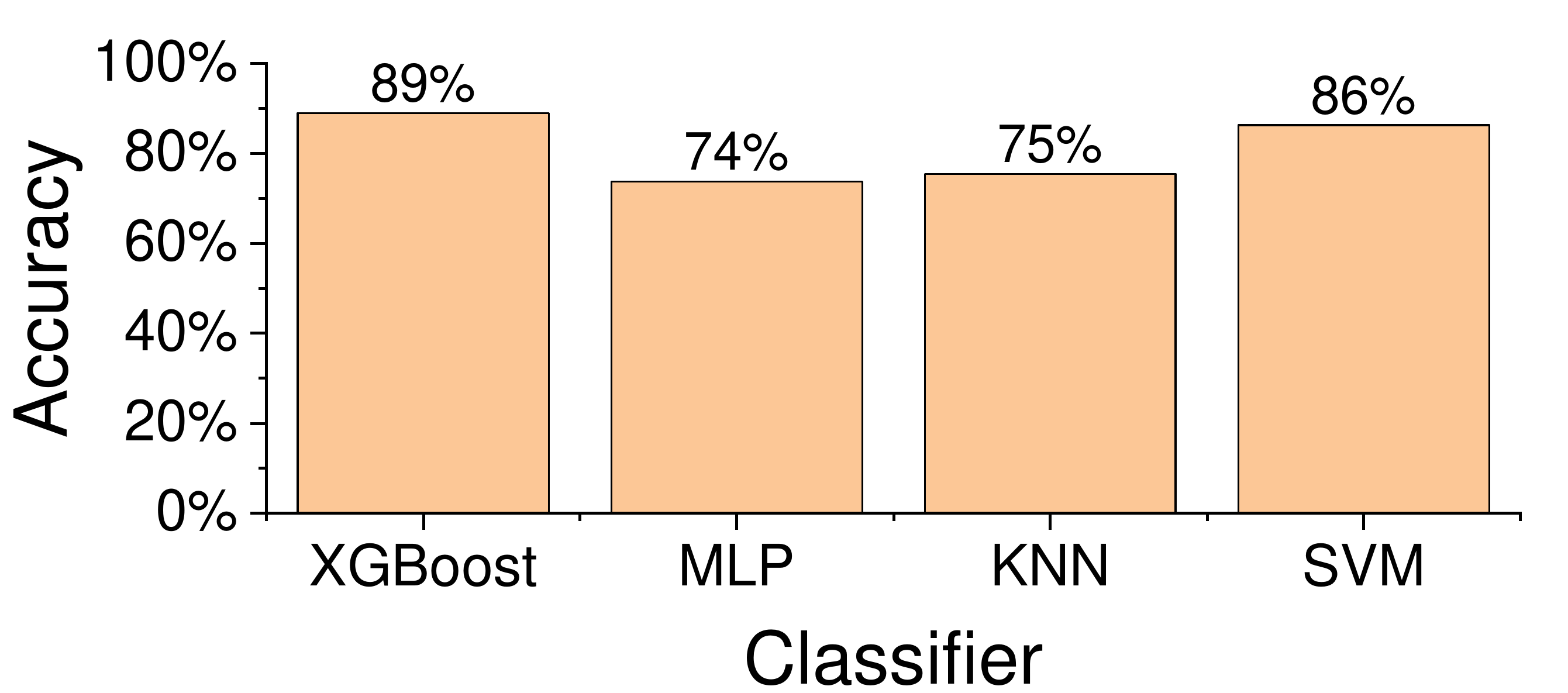}\\
    \end{minipage}
}
\subfigure[Inference Time] { \label{fig:inferenceTime}
    \begin{minipage}[t]{0.5\linewidth}
        \centering
        \includegraphics[width=1\textwidth]{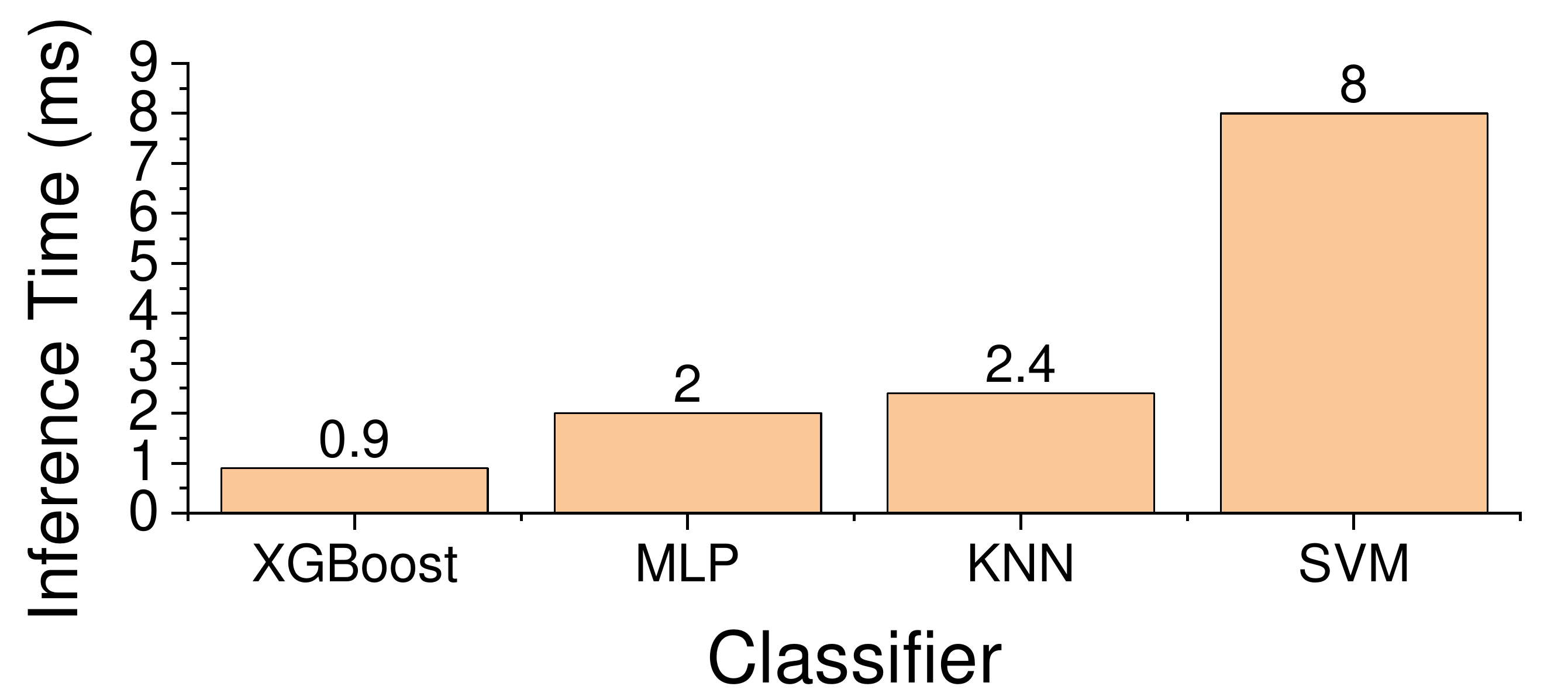}\\
        \vspace{0.2cm}
    \end{minipage}%
}%
\centering
\vspace{-4mm}
\caption{Comparing our XGBoost model against alternative modeling techniques.}
\label{fig:alternatemodel}
\vspace{-5mm}
\end{figure}

\cparagraph{Alternative modeling techniques.} Figure \ref{fig:alternatemodel} compares our XGBoost-based predictor against three other
classification methods used in prior works for code optimization~\cite{wang2018machine}:  MLP neural network \cite{gardner1998artificial},
KNN (with $k=1$) \cite{zhang2007ml}, and SVM \cite{noble2006support}. All the alternative techniques were trained and evaluated using the
same method and training data as our model. In this experiment, we consider the model prediction accuracy and the time for making a
prediction. As can be seen from the diagram, our approach has the lowest runtime overhead while giving the highest accuracy when compared
to alternative modeling techniques. Since XGBoost is a decision-tree-based model, it also has the advantage of being interpretable because
its decision process can be followed by traversing the tree.

\cparagraph{Training and deployment overhead.} Training of our predictive model only needs to be performed once, after which the trained model can be applied to any matrices. Training is dominated by the generation of training data which takes in total less than a week’s machine time (Section \ref{sec:dg}). We can speed this up by using multiple machines. The overhead for learning the XGBoost model is negligible, less than 5 minutes.
 Our approach has a negligible runtime overhead compared to the GNN kernel execution time, the overhead of feature extraction and prediction is less than 3\% to the end-to-end kernel execution time.

\vspace{-4mm}
\subsection{Discussion}
\vspace{-3mm}

\cparagraph{Supporting other storage formats.} Our approach can be easily extended to support other sparse matrix storage formats. As we
formulate the storage format prediction as a classification problem, this can be achieved by adding a new class label (for the newly
supported format) into our training dataset. Doing so would also require providing the relevant SpMM kernel implementation. Other than
these, a large part of the training process and deployment can remain unchanged.

\cparagraph{Supporting GPU computation}. This work focuses on the CPU execution of GNN models due to the large graph datasets that a GNN
model typically processes. There are methods to support large-scale graph processing on GPUs such as GraphSAGE
\cite{hamilton2017inductive}. Our approach can be ported to support GPU processing. This will require using training data collected from
the targeting GPU to train our predictive model. 

\cparagraph{Optimize SpMM algorithms.} Optimizing SpMM computation is an active research field \cite{dalton2015optimizing}. It is interesting to investigate how the SpMM computation kernel can be tailored for GNN computation and what parameters can be opened to a tuning framework. As the best algorithm parameters are likely to change depending on the matrix input and the underlying hardware, an automatic machine learning-based approach similar to our approach is highly attractive.

\vspace{-5mm}
\section{Related Work}
\vspace{-4mm}

Several approaches have been proposed to optimize graph processing \cite{xie2020gnns}. Some provide new programming abstractions to
optimize vertex/node-centric or edge-centric  processing \cite{zhou2020graph}. For example, Pytorch-Geometric (PyG)
\cite{fey2019fast} and Deep Graph Library (DGL) \cite{wang2019deep} are two major frameworks for GNN computation. Both libraries rely on a
low-level, hand-optimized SpMM library, but they use a single sparse matrix storage format throughout the execution. 
Our work complements these prior efforts by dynamically adapting the sparse matrix storage format and the associated computation kernel for
each GNN layer, which can be easily integrated with existing graph programming models.

Various sparse matrix storage formats have been proposed in the past \cite{langr2015evaluation}. Studies have shown
that there is no ``one-fit-for-all" storage format, and the right format can change from one matrix to the other
\cite{li2013smat,chen2020characterizing}. Methods have been proposed to dynamically choose sparse matrix storage format based on the input
workloads \cite{sedaghati2015automatic}. These include approaches build around analytical methods \cite{venkat2015loop} or
machine-learning-based predictive models \cite{chen2019optimizing}. The latter has the benefit of can be easily ported to different
architectures as machine learning learns from empirical observations rather than simplified assumptions used by an analytical model.
However, prior machine-learning-based solutions have been concentrated on optimizing sparse matrix-vector multiplication (SpMV) of
scientific workloads \cite{zhao2018bridging}. They choose a storage format at the beginning of the program execution but do not adjust the
format during application execution. No work so far has concerned choosing the sparse matrix storage format for GNN SpMM throughout program
execution. Our work is the first to do so.

Machine learning is a proven design methodology for systems modeling and optimization \cite{wang2009mapping,ren2017optimise,wang2018machine,zhang2018auto,wang2014integrating,zhang2020optimizing}. Studies have demonstrated
the success of applying machine learning for a wide range of code optimization tasks
\cite{tournavitis2009towards,wang2014automatic,cummins2017end,wang2010partitioning,ye2020deep,wang2020combining} In this work,
we employ machine learning techniques to develop an automatic approach to optimize GNN SpMM. We remark that our work does not seek to
advance machine learning algorithms; instead, it explores and applies a well-established modeling method to tackle the GNN SpMM
optimization problem.

\vspace{-7mm}
\section{Conclusions}
\vspace{-4mm} This paper has presented a machine-learning based predictive model to dynamically choose the sparse matrix storage format and
the associate computation kernel during GNN execution. Our model uses numerical features to characterize the input matrix to predict
the storage format to use for the next GNN layer. We evaluate our approach by applying it to five representative GNN models
running on a multi-core CPU using both real-world and synthetic datasets. Experimental results show that our approach gives an average
speedup of 1.17x (up to 3x) over the Pytorch default strategy and exhibits a good generalization ability.

\vspace{-5mm}
\bibliographystyle{splncs04}
\bibliography{refs}
\end{document}